\PassOptionsToPackage{unicode=true}{hyperref}
\RequirePackage{nameref}
\documentclass[electronics arxiv,article,submit,pdftex,moreauthors]{Definitions/mdpi} 
\usepackage[T1]{fontenc}
\usepackage{lmodern}
\usepackage{algorithm, algorithmicx, algpseudocode}
\usepackage{xcolor,mathtools, comment}
\usepackage{graphicx}
\usepackage{listings}
\renewcommand{\linenumbers}{}
\firstpage{1} 
\pubvolume{1}
\issuenum{1}
\articlenumber{0}
\pubyear{2022}
\copyrightyear{2022}
\datereceived{} 
\dateaccepted{} 
\datepublished{} 
\hreflink{https://doi.org/} % If needed use \linebreak
\Title{DVA-Neurons: Design and Verification of Adaptive LIF Neurons: From Single-Neuron Dynamics to Multi-Neuron Spiking Networks}

\TitleCitation{Title}

\Author{Thanh Pham $^{1}$ and Riadul Islam $^{2,\dagger}$\orcidA{}} %, Firstname Lastname $^{2,\ddagger}$ and Firstname Lastname $^{2,}$*}
\AuthorNames{Firstname Lastname, Firstname Lastname and Firstname Lastname}

\AuthorCitation{Lastname, F.; Lastname, F.; Lastname, F.}
\address{%
$^{1}$ \quad Department of Computer Science and Electrical Engineering, University of Maryland, Baltimore County, MD 21250, United States; thanhp1@umbc.edu\\
$^{2}$ \quad Department of Computer Science and Electrical Engineering, University of Maryland, Baltimore County, MD 21250, United States; riaduli@umbc.edu}
\corres{Correspondence: riaduli@umbc.edu} %; Tel.: (optional; include country code; if there are multiple corresponding authors, add author initials) +xx-xxxx-xxx-xxxx (F.L.)}

\abstract{Spiking Neural Networks (SNNs) offer a promising path toward ultra-low-power artificial intelligence inference by emulating the event-driven computation of biological neurons. However, two challenges limit their practical deployment. First, fixed-parameter Leaky Integrate-and-Fire (LIF) neurons lack the adaptation mechanisms observed in biology, where neurons modulate their excitability based on firing history. Second, scaling from single neurons to multi-neuron networks introduces challenges in synaptic weight distribution and inter-neuron spike routing that are absent in isolated designs. This paper addresses both issues through the extension, verification, and physical implementation of adaptive LIF neurons at three architectural scales. Building on Lin's 1st-order adaptive LIF neuron — previously developed as a Tiny Tapeout macro — this work contributes: a 2nd-order neuron with two-stage synaptic filtering for richer temporal dynamics; a fully-connected 6-neuron spiking network with configurable weights (100 to 5) demonstrating weight-based inter-neuron communication; and a direct verification methodology enabling per-cycle observation of all internal states. All designs were synthesized targeting Selected Area Electron Diffraction (SAED) 14 nm Complementary Metal-Oxide-Semiconductor (CMOS) technology at 1 GHz and verified with Cocotb-based Python testbenches under pulsed current stimuli (amplitude 80, ISI=3). The results show that adaptation effectively modulates firing: 31\% suppression in the 2nd-order neuron (25 vs.\ 36 spikes) and 31\% reduction in postsynaptic firing in the network (18 vs.\ 26 spikes). Physically, the 2nd-order neuron costs 1.77$\times$ more area and 1.52$\times$ more power than the 1st-order baseline, while the 6-neuron network demonstrates near-linear scaling (5.7$\times$ area, 5.3$\times$ power). Seven verification bugs spanning testbench connectivity, fixed-point overflow, and Verilog expression-width semantics are documented.}

\keyword{Spiking Neural Networks; Adaptive LIF Neurons; Neuromorphic Computing; Hardware Verification.} 

\begin{document}

\section{Introduction}
\label{sec:intro}

The energy efficiency~\cite{venkataramani2014axnn, Islam_tcasii:2021, islam2011high} %~\cite{Islam_tcasii:2021}
 gap between biological brains and conventional computing architectures has motivated decades of research in neuromorphic engineering. Although modern deep learning frameworks can achieve hundreds of tera FLOPs per second, the power consumption is exponentially higher than the 20~W required by the human brain~\cite{Fedorova:2024}. SNNs address this gap by representing information as discrete, asynchronous electrical spikes that are analogous to biological action potentials~\cite{Huo:2025, Islam_benchmarking:2024}. Because computation is event-driven, SNN hardware~\cite{zhang2020low, javanshir2022advancements, han2020hardware, song2020compiling} consumes negligible power during idle periods, making it well-suited for always-on edge sensing applications in the Internet of Things (IoT)~\cite{Du_snn-iot:2025, KIM2026107953, niu2026spikegate, bao2026tpipe} and event-based computation or vision platforms~\cite{cheng2025fpga, wei2024event, Islam_isvlsi:2025, shahsavari2023advancements, Islam_facedataset:2024, mule2025descriptor, blouw2020event}.

%{\color{blue}
To realize neuromorphic processing in physical hardware, researchers have explored a wide range of physical devices. Specifically, this hardware falls into two categories: standard CMOS integrated circuits and emerging devices such as memristors~\cite{ji2025ultralow}, magnetic tunnel junctions (MTJs)~\cite{zhu2025two}, floating-gate transistors~\cite{danial2019two, robucci2010compressive}, and ferroelectric FETs~\cite{khan2020future, islam2018negative}. On the CMOS side, in-memory computation~\cite{ko2025cmos, challagundla2024arxrcim, niu2026all, Challagundla_iccd:2023, challagundla2025tdc} , current-mode systems~\cite{Islam_DCM:2015, islam2018dcmcs}, resonant energy recycling techniques~\cite{islam2011high, zou2025efficient, challagundla2023design, challagundla2022power}, and event-based vision~\cite{Lu_2026_CVPR, islam_event:2026} are attracting significant attention because of their ultra-low-power operation. On the developing material side, emerging technologies -- including non-volatile memristive arrays, oxide-based thin-film transistors, and multi-terminal optoelectronic synaptic devices -- emulate biological plasticity natively through physical state changes~\cite{Hu2021OptoelectronicNC}. These emerging devices offer high integration density, analog weight tuning, and low-energy state transitions directly at the material level by leveraging light, electrical pulses, or ion migration to modulate conductance~\cite{Hu2021OptoelectronicNC}. However, despite their promise for analog in-memory computing, emerging devices often face challenges regarding device-to-device variation, manufacturing yield, and complex integration with peripheral control logic~\cite{Hu2021OptoelectronicNC}.

Besides, digital CMOS integrated circuits remain the primary foundation for scalable, short-term deployed neuromorphic hardware due to their high noise margins, deterministic execution, and compatibility with mature commercial foundry processes. Large-scale digital processors like Intel’s Loihi~\cite{Davies:Loihi2018} and IBM’s TrueNorth~\cite{Sawada_IBM:2016} demonstrate the viability of this approach. Within digital CMOS implementations, the Leaky Integrate-and-Fire (LIF) model is the most widely used abstraction of neuronal dynamics in neuromorphic hardware~\cite{Lapicque:1907, lapicque_trans:2007, Stein:1967}. It captures the essential neuronal behaviors of charge integration, passive membrane leakage, threshold-triggered firing, and post-spike reset while remaining computationally efficient for large-scale implementations~\cite{Huo:2025}. There are also some studies on adapting the Spiking Neuron Network to Deep Neural Network with some promising results~\cite{LuLIFModel:2022}. However, the classical fixed-parameter LIF model does not capture the adaptive mechanisms observed in biological neurons, where excitability changes according to firing history~\cite{yang2025spiking, wang2023complex, CHENG2023217, zhu2024spikegptgenerativepretrainedlanguage}. Furthermore, extending from individual neurons to multi-neuron networks introduces additional challenges, including synaptic weight distribution and spike-based communication among neurons~\cite{Islam_icrest:2023, kang2025npeva, karthikeyan20253d, zhang2026pipelined}.

This work builds on Lin's first-order adaptive LIF neuron~\cite{lin:25}, which introduced adaptive threshold and adaptive beta mechanisms. The contributions of this work are threefold. 
\begin{itemize}
    \item First, a second-order LIF neuron is developed by introducing the synaptic current state variable with independent decay kinetics ($\alpha = 205/256$), enabling richer temporal dynamics while quantifying the hardware overhead of the additional state variable. 
    \item Second, a fully connected six-neuron network is implemented, in which one input neuron (N0) drives five postsynaptic neurons (N1--N5) through software--configurable synaptic weights of 100, 30, 15, 10, and 5, demonstrating spike-based communication among independently adapting second-order LIF neurons. 
    \item Third, a direct-interface verification methodology replaces the Tiny Tapeout I/O wrapper with cycle-by-cycle monitoring of internal states, allowing simultaneous observation of membrane potential, synaptic current, adaptive threshold, adaptive beta, and all six neuron spike outputs. All three designs are verified using Cocotb-based Python testbenches and synthesized through a complete ASIC flow targeting the SAED 14~nm technology at 1~GHz. %The complete Verilog source code, testbenches, and synthesis scripts are available in the project repository. We can add this sentence once we have a public repo.
\end{itemize}

%%%%%%%%%%%%%%%%%%%            BACKGROUND SECTION           %%%%%%%%%%%%%%%%%%%%
\section{Background}
\label{sec:background}

\subsection{The Leaky Integrate-and-Fire Neuron Model}

The LIF neuron model describes a point neuron whose membrane potential $V(t)$ changes according to a first-order differential equation. Input current $I(t)$ charges the membrane capacitance, while a leakage conductance continuously drives the potential back toward a resting value. When $V(t)$ crosses a threshold $V_{th}$, the neuron emits a spike, the potential is reset to zero, and a refractory period begins. In discrete-time digital implementation, this becomes a difference equation:
\begin{equation}
V[t+1] = I[t] + V[t] \cdot \beta
\end{equation}

where $\beta < 1$ encodes the leakage rate. This 1st-order formulation uses a single state variable and produces regular, predictable firing patterns under constant input.

\subsection{Synaptic Dynamics and 2nd-Order Models}

In biological neurons, incoming spikes trigger neurotransmitter release that opens ion channels on the postsynaptic membrane, generating a synaptic current with its own temporal dynamics. A 2nd-order LIF model captures this by introducing an explicit synaptic current variable $S[t]$ that filters the input before driving the membrane:

\begin{equation}
S[t+1] = I[t] + S[t] \cdot \alpha  \quad  \text{(synaptic filtering)}
\end{equation}

\begin{equation}
V[t+1] = S[t+1] + V[t] \cdot \beta  \quad \text{(membrane integration)}
\end{equation}

\begin{figure}[tb]
   \centering
   \includegraphics[width=1\columnwidth]{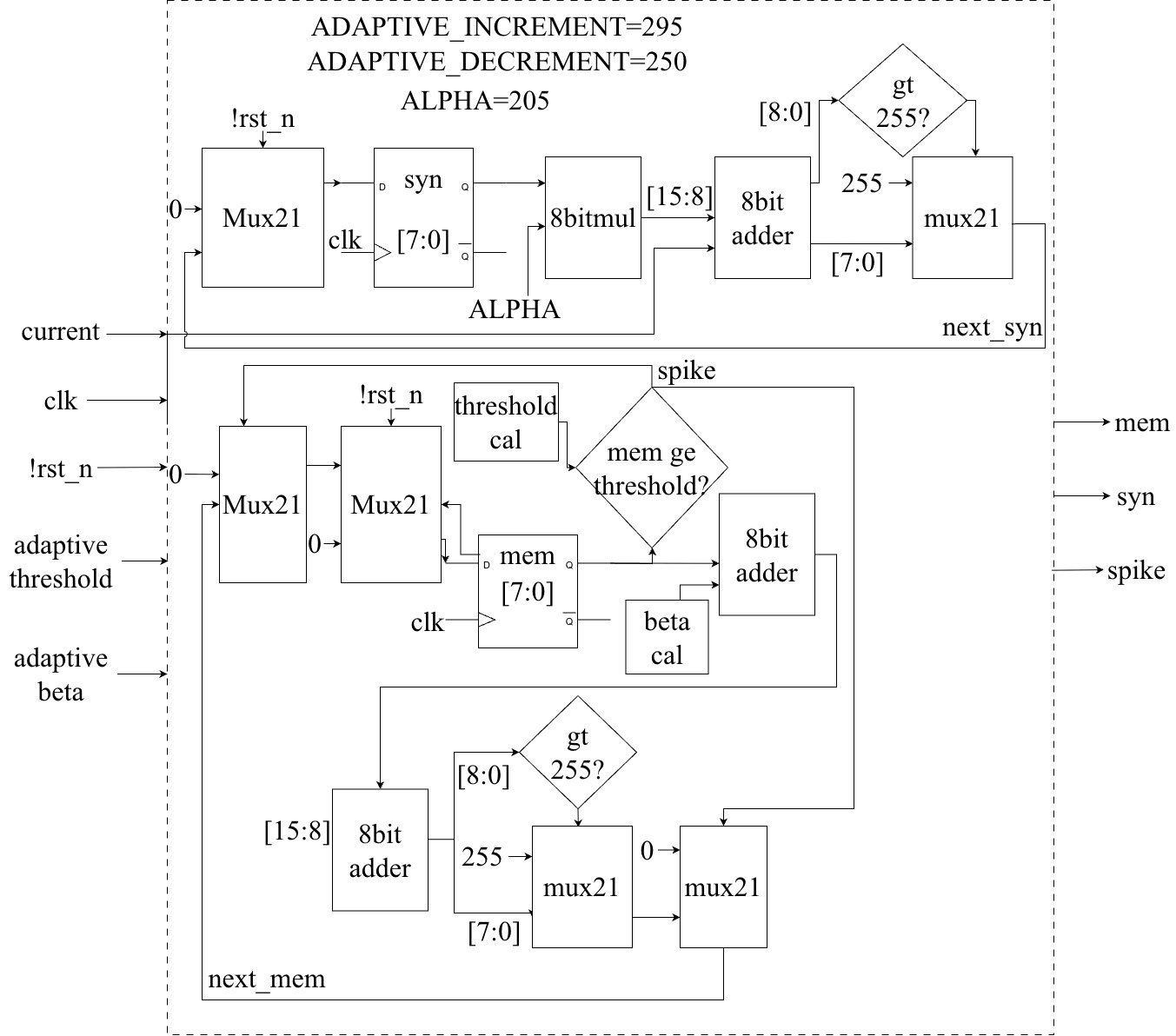}
   \caption{Block diagram of the 2nd-order LIF neuron showing the two-stage pipeline: synaptic filtering ($\alpha$) followed by membrane integration ($\beta$), with adaptive threshold and adaptive beta feedback paths.}
   \label{fig:lif_2nd_block}
\end{figure}

Here, $\alpha$ controls the synaptic decay rate. This two-stage cascade introduces short-term memory: the synapse accumulates input over time, smoothing transient fluctuations and producing more complex firing patterns. This concept of leveraging prior timestep information aligns with recent work by Xia et al. \cite{XIA2025130253}, whose CPT-SNN architecture explicitly combines previous timestep states to improve temporal representation in spiking networks. The biological realism comes at a hardware cost — a second multiply-accumulate path, additional registers, and a wider datapath. In addition to the short-term memory characteristic, the synapse adaptation can also mimic the long-term memory feature by implementing the synapse weight adaptation~\cite{Carlson:2013}. By learning the synapse weight every time step, the neurons can benefit from storing the important synapse weight corresponding to significant inputs. However, this paper focuses mainly on short-term memory to prove the plausibility of hardware implementation, so long-term memory can be the next goal in future work. Fig.~\ref{fig:lif_2nd_block} illustrates the two-stage architecture.

\subsection{Adaptive Mechanisms}

\begin{figure}[tb]
   \centering
   \includegraphics[width=1\columnwidth]{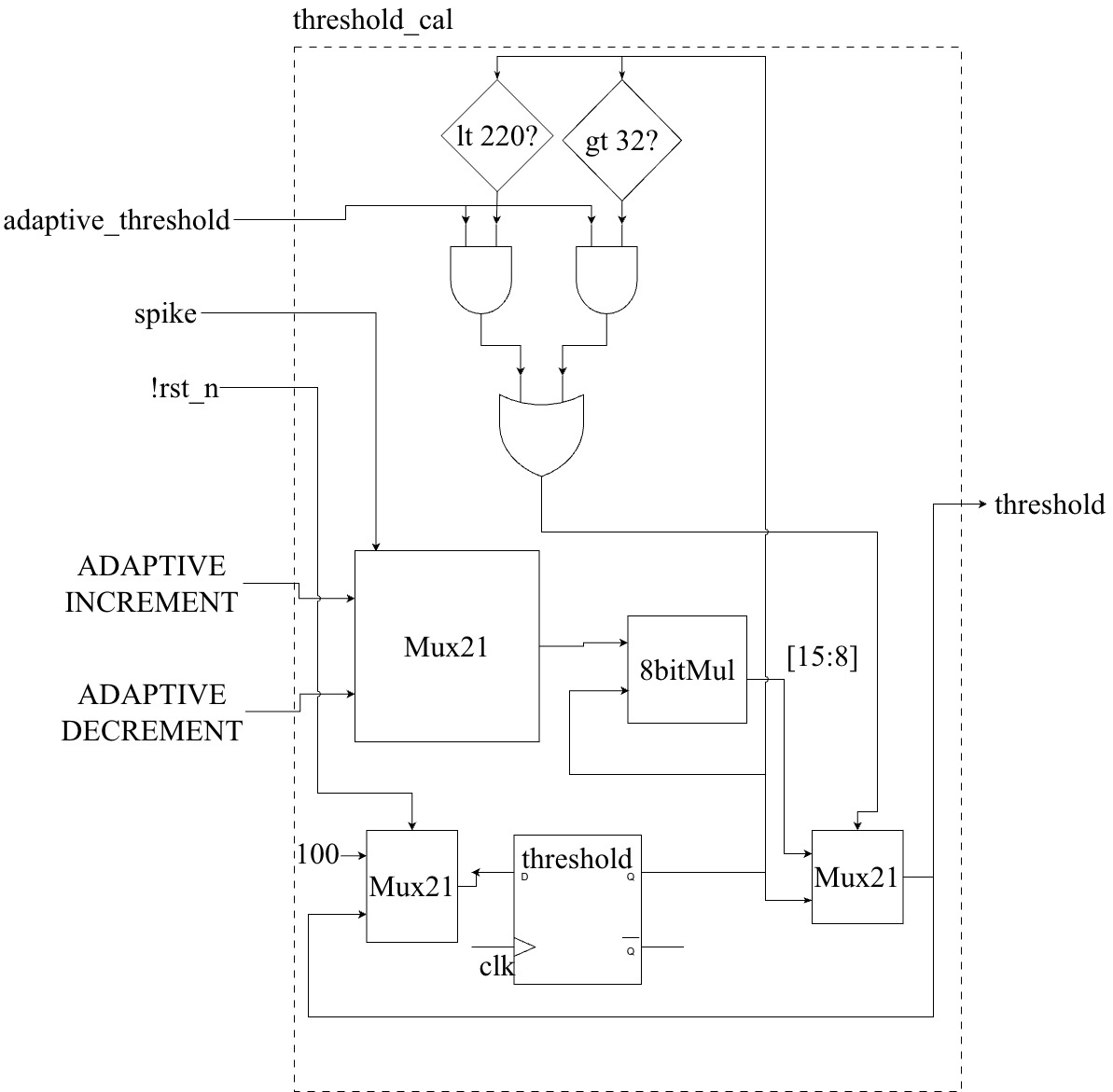}
   \caption{Adaptive threshold calculation. On spike: $threshold \leftarrow threshold \times 295/256$ (increase, clamped to 220). On non-spike: $threshold \leftarrow threshold \times 250/256$ (decay, clamped to 32).}
   \label{fig:threshold_cal}
\end{figure}

\begin{figure}[tb]
   \centering
   \includegraphics[width=0.85\columnwidth]{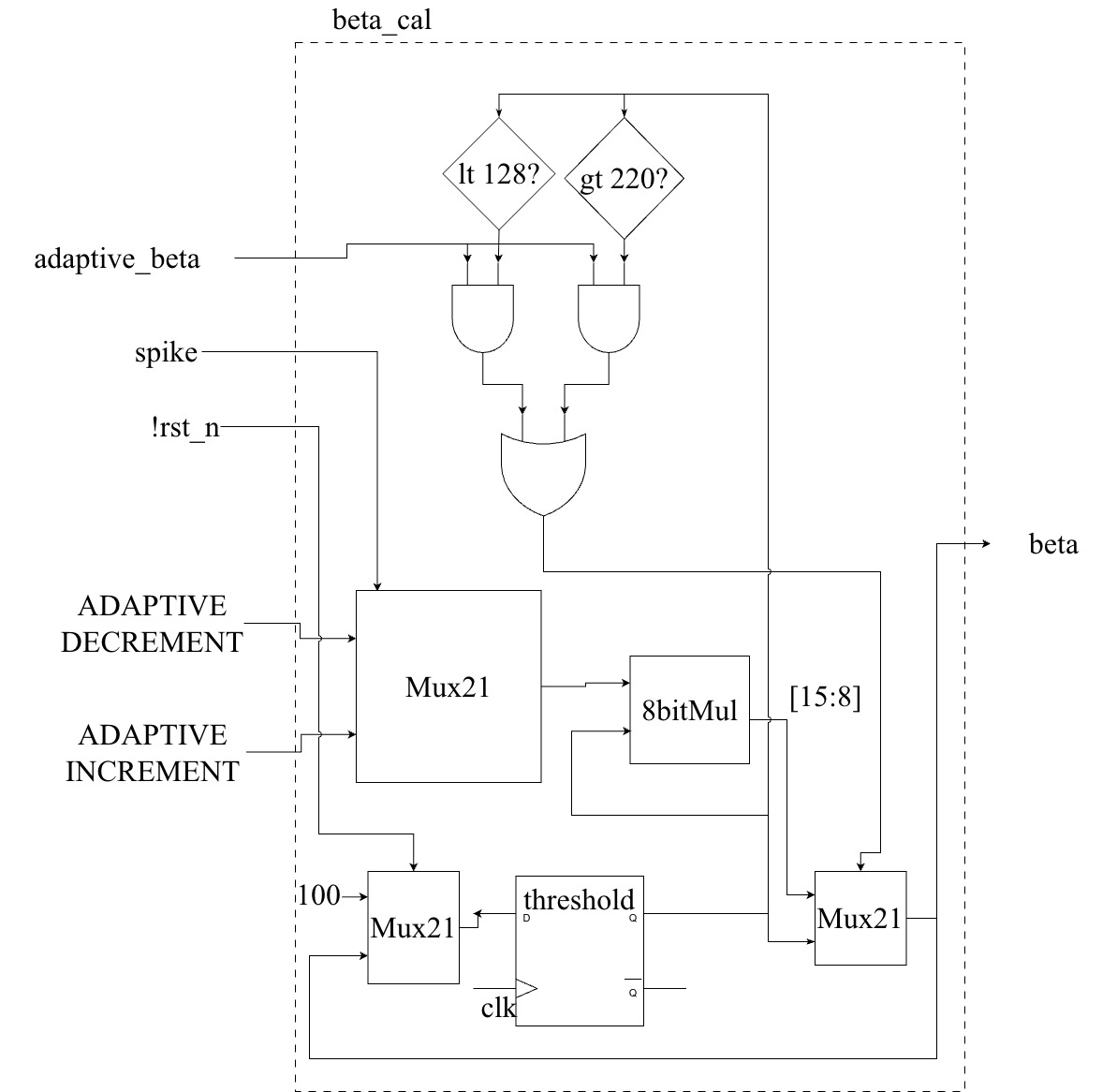}
   \caption{Adaptive beta calculation. On spike: $\beta \leftarrow \beta \times 250/256$ (more leak, clamped to 128). On non-spike: $\beta \leftarrow \beta \times 295/256$ (less leak, clamped to 220).}
   \label{fig:beta_cal}
\end{figure}

Cortical neurons exhibit spike-frequency adaptation: their firing rate decreases under sustained stimulation as after-hyperpolarization currents build up ~\cite{Carlson:2013}. Two complementary adaptation mechanisms in digital logic are implemented. First, the firing threshold increases after each spike and decays during quiet periods, making the neuron progressively harder to excite during high-activity epochs. Second, the membrane decay parameter $\beta$ decreases after spiking (more leak, faster return to rest) and recovers during silence (less leak, greater temporal integration). By keeping the $threshold$ in the range from 32 to 220 and $\beta$ in the range from 128 to 220, these prevent both runaway firing and complete silence~\cite{Carlson:2013}. Fig.~\ref{fig:threshold_cal} and Fig.~\ref{fig:beta_cal} detail the update logic.

\subsection{Multi-Neuron Network Architecture}

To evaluate inter-neuron spike communication, a 6-neuron fully-connected network was implemented. A single input neuron (N0) receives external pulsed current and generates spike events. Its spike output fans out to five postsynaptic neurons (N1--N5) through configurable weight parameters. When N0 fires, each layer-1 neuron receives a current injection equal to its weight on that cycle:

\begin{equation}
I_k[t] = spike_0[t] \cdot W_k, \quad k \in \{1..5\}
\end{equation}

with weights $W_1=100$, $W_2=30$, $W_3=15$, $W_4=10$, $W_5=5$. Each of the six neurons is an independent 2nd-order LIF instance with its own membrane, synapse, threshold, and beta registers. A neuron select input allows the testbench to read any neuron's membrane and synapse values individually, while a broadcast mode simultaneously exposes all six spike signals on the output bus, enabling real-time observation of the entire network's spiking activity. The architecture provides a testbed for studying weight-dependent synaptic transmission, rate coding, and adaptive communication in a small-scale SNN~\cite{Huo:2025}.

\subsection{Fixed-Point Arithmetic for Digital Neurons}

Implementing neural dynamics in digital hardware requires discretization of continuous parameters. 8-bit fixed-point arithmetic with implicit scaling is implemented by 1/256, where multiplication by a parameter $P$ is implemented as integer multiplication followed by an 8-bit right shift. For example, $\alpha = 205/256$ approximates 0.8, and an initial $\beta$ of $128/256$ approximates 0.5. This representation eliminates floating-point units, reducing area and power at the cost of quantization error. The 8-bit dynamic range (0--255) constrains maximum membrane potential and introduces risk of arithmetic overflow in the two-stage integration path, a challenge addressed in the methodology.

\subsection{Related Neuromorphic Hardware}

Intel's Loihi and IBM's TrueNorth represent the state of the art in large-scale neuromorphic processors, demonstrating 100--1000$\times$ energy efficiency improvements over conventional architectures for spiking neural network workloads~\cite{Merolla2014AMS, Intel2021Neuromorphic, Davies:Loihi2018}. These chips implement programmable LIF neuron models alongside on-chip learning rules such as spike-timing-dependent plasticity (STDP), a topic surveyed comprehensively by Rahman and Yusoff \cite{RAHMAN2025129170}. Huo et al. \cite{Huo:2025} provide a recent review of biologically-plausible SNN learning algorithms. The Tiny Tapeout platform and Skywater 130nm open-source PDK have recently democratized ASIC design \cite{lin:25}. Our work builds on these foundations by providing a systematic comparison across three architectural scales through a complete physical design flow, with detailed documentation of verification challenges unique to fixed-point neural dynamics.

At the individual neuron level, digital SNN accelerators balance datapath width, adaptive feature complexity, area, and clock frequency. Lin’s TT05 demonstration introduced an 8-bit 1st-order adaptive LIF neuron targeting the SkyWater 130 nm open-source PDK~\cite{lin:25}. This work expands upon that foundation by evaluating both 1st-order and proposed 2nd-order adaptive LIF designs in a high-performance 14 nm FinFET process operating at $1\text{ GHz}$.

Table~\ref{tab:neuromorphic_comparison} provides a quantitative comparison between published digital neuromorphic architectures and the designs presented in this work.

\begin{table*}[tb]
\caption{Quantitative Comparison of Digital Neuron Architectures and Neuromorphic Processors}
\label{tab:neuromorphic_comparison}
\small
\renewcommand{\arraystretch}{1.3} % Adjust row height (1.3 to 1.5 is ideal)
\begin{center}
\begin{tabular}{|l|l|c|c|}
\hline
\textbf{Architecture/Work} & \begin{tabular}{@{}c@{}}\textbf{Dynamic Adaptation}\\\textbf{Features}\end{tabular} & \begin{tabular}{@{}c@{}}\textbf{Core Area/}\\\textbf{Neuron($\mu\text{m}^2$)}\end{tabular} & \begin{tabular}{@{}c@{}}\textbf{Dynamic Power/}\\\textbf{Neuron($\mu\text{W}$)}\end{tabular} \\
\hline
IBM TrueNorth & None (Fixed LIF) & $\sim 365$ & $0.02$ \\
Intel Loihi & \begin{tabular}{@{}l@{}}Programmable Threshold \&\\STDP\end{tabular} & $\sim 390$ (amortized) & $\sim 20\text{--}50$ \\
TT05 LIF Baseline & Adaptive Threshold \& $\beta$ & N/A & N/A \\
\textbf{This 1st-Order} & Adaptive Threshold \& $\beta$ & \textbf{303.56} & $\mathbf{83.60}$ \\
\textbf{This 2nd-Order} & \begin{tabular}{@{}l@{}}2nd-Order Synaptic +\\Adapt Thresh/$\beta$\end{tabular} & \textbf{393.78} & $\mathbf{127.00}$ \\
\textbf{This 6-Neuron Net} & \begin{tabular}{@{}l@{}}Fully Connected +\\Multi-Neuron Adapt\end{tabular} & \begin{tabular}{@{}c@{}}\textbf{374.02}\\(avg/neuron)\end{tabular} & \begin{tabular}{@{}c@{}}\textbf{111.50}\\(avg/neuron)\end{tabular} \\
\hline
\end{tabular}
\end{center}
\end{table*}

%%%%%%%%%%%%%%%%%%%           PROPOSED DESIGN SECTION          %%%%%%%%%%%%%%%%%%%%
\section{Methodologies}
\label{sec:methodologies}

\subsection{Design Architecture and Test Interface}

All three designs use a direct Verilog module interface with independently accessible ports: an 8-bit \texttt{current} input for stimulus, separate enable lines for \texttt{adaptive\_threshold} and \texttt{adaptive\_beta}, and dedicated output ports for \texttt{state}/\texttt{mem}, \texttt{syn} (2nd-order), and \texttt{spike}. Unlike Tiny Tapeout-compatible wrappers that pack signals through multiplexed 8-bit I/O buses, the direct interface exposes every internal state signal for simultaneous observation in simulation. The 6-neuron network additionally provides a \texttt{neuron\_select} port for reading individual neuron membrane and synapse values, and an \texttt{all\_spikes[5:0]} broadcast port showing all six spike signals simultaneously. This eliminates the need for cycle-by-cycle register poking to extract the full network state.

\subsection{1st-Order Neuron Design (Baseline)}

The 1st-order LIF neuron (lif.v), originally designed by Lin \cite{lin:25}, serves as the baseline architecture. It maintains a single 8-bit membrane state register and two adaptive parameter registers (threshold and beta). The next-state equation is:

\begin{equation}
    next\_state = current + (state \cdot \beta) \gg 8
\end{equation}

where the spike condition gates both the input current term and the decay term to zero on reset, and $\gg$ denotes logical right shift. The sequential logic block updates threshold and beta at each positive clock edge: threshold multiplies by $295/256$ on spike (increase) and by $250/256$ on non-spike (decay), while beta follows the inverse pattern. Both parameters are clamped to prevent overflow and underflow. The initial threshold is 100 (mid-range), and the initial beta is $224$ ($\sim0.875$, slow decay).

\subsection{Proposed 2nd-Order Neuron Design}

The 2nd-order neuron (lif\_2nd.v) extends the baseline architecture by introducing a second 8-bit state register for synaptic current alongside the membrane potential register, plus identical threshold and beta adaptive registers. The two-stage discrete-time update is:

\begin{equation}
    syn[t+1] = \alpha \cdot syn[t] + current[t]
\end{equation}

\begin{equation}
    mem[t+1] = \beta \cdot mem[t] + syn[t+1]   \quad  \text{(reset to 0 if spike)}
\end{equation}

with $\alpha = 205/256$ ($\approx 0.8$) implemented as a fixed parameter and $\beta$ initialized to $128/256$ ($\approx 0.5$, faster leak than the 1st-order). A critical implementation detail concerns Verilog's expression width rules: the product of two 8-bit operands is evaluated in the context of the assignment target width. If $syn \cdot \alpha$ is directly used in an expression whose result feeds an 8-bit wire, Verilog truncates the product to 8 bits before the right shift, making $(syn \cdot \alpha) \gg 8$ identically zero for any $syn < 256$. To prevent this, both multiply operations use explicit 16-bit wire declarations ($syn\_product$ and $mem\_product$) that force the synthesis tool to preserve the full 16-bit product before shifting and reduction. This was the single most impactful bug encountered during development and is discussed in the results.

Overflow prevention is handled through 9-bit saturation comparators. The intermediate sums $next\_syn\_full$ and $next\_mem\_full$ are computed at 9-bit width. If the sum exceeds 255, the saturated value is clamped to 255. This prevents wraparound corruption in the 8-bit datapath.

The synaptic current register updates unconditionally every cycle — regardless of whether the neuron spikes — reflecting the biological reality that synaptic conductance dynamics is independent of postsynaptic firing. The membrane potential resets to zero on spike. This design decision (bug fix \#5) was essential for correct 2nd-order behavior, as the original code only updated $syn$ during non-spike cycles, causing the synaptic state to stagnate at zero during high-frequency firing. Fig.~\ref{fig:verilog_code} shows the key Verilog implementation excerpt.

\begin{figure}[tb]
   \centering
   \begin{lstlisting}[basicstyle=\ttfamily\scriptsize, frame=single, language=Verilog]
// 16-bit wires to prevent 8-bit truncation
wire [15:0] syn_product = syn * ALPHA;
wire [15:0] mem_product = mem * beta;
// 9-bit sums with saturation
wire [8:0] next_syn_full = current + (syn_product >> 8);
assign next_syn = (next_syn_full > 8'd255) 
? (syn_product >> 8) : next_syn_full[7:0];
wire [8:0] next_mem_full = next_syn + (mem_product >> 8);
assign next_mem = spike ? 8'd0 : 
((next_mem_full > 8'd255) ? 8'd255 : next_mem_full[7:0]);
   \end{lstlisting}
   \caption{Verilog excerpt of the 2nd-order LIF neuron showing the 16-bit product wires, 9-bit saturation comparators, and the unconditional synaptic current update.}
   \label{fig:verilog_code}
\end{figure}

\subsection{Network Design}

The 6-neuron network (lif\_network.v), a new contribution of this work, instantiates six 2nd-order LIF neurons. N0 receives the external \texttt{current} input directly. N1--N5 each receive a gated current: when N0 fires ($spike_0 = 1$), the neuron's weight is injected as a one-cycle current pulse; otherwise, zero current is applied. The weights are compile-time parameters (100, 30, 15, 10, 5) providing a 20:1 dynamic range from strongest to weakest synapse. All six neurons share common \texttt{adaptive\_threshold} and \texttt{adaptive\_beta} control lines, enabling consistent adaptation policies across the network. The \texttt{neuron\_select} port and \texttt{all\_spikes} broadcast output allow the testbench to observe the full network state cycle by cycle.

\subsection{Verification Methodology}

Functional verification was performed using Cocotb, a Python-based coroutine testbench framework, driving Icarus Verilog simulations. Each testbench operates at a 1 ns clock period (1 GHz effective rate) and reads the device state on every clock cycle — unlike earlier Tiny Tapeout-wrapped testbenches that could only sample at the end of each pulse train. The per-cycle read approach reliably captures all spike events regardless of which phase (pulse, quiet-1, quiet-2) they fire on.

All three designs are tested with five behavioral scenarios under pulsed current drive: each pulse delivers amplitude 80 for one cycle followed by a baseline of 6 for the remaining gap cycles:

\begin{enumerate}
    \item \textbf{Both adaptive mechanisms enabled} at ISI=3 (25 pulses) — evaluates the interaction of threshold and beta adaptation under periodic excitation.
    \item \textbf{Adaptive threshold only} at ISI=3 (25 pulses) — isolates threshold adaptation dynamics.
    \item \textbf{Adaptive beta only} at ISI=3 (25 pulses) — isolates decay rate adaptation.
    \item \textbf{Both mechanisms disabled} at ISI=3 (25 pulses) — establishes the pure fixed-parameter LIF baseline.
    \item \textbf{Rate coding} with no adaptation — verifies that firing rate tracks stimulus frequency at slow (ISI=4, 10 pulses), medium (ISI=2, 15 pulses), and fast (ISI=1, 20 pulses) rates.
\end{enumerate}

For the network, additional scenarios test one-to-one spike communication (N0 to N1), fan-out to all five receivers, strong-vs-weak synapse comparison (N1 with W=100 vs N5 with W=5), and adaptive communication across N0$\rightarrow$N1.

\subsection{Physical Design Flow}

All three designs were carried through a common physical implementation flow targeting the SAED 14 nm educational process design kit at 0.8V nominal supply, slow-slow (SS) corner. Logic synthesis was performed with Synopsys Design Compiler using $compile\_ultra$ with a $1 ns$ clock period constraint. The resulting gate-level netlists were passed to Synopsys ICC2 for physical design: floorplanning at $60\%$ core utilization with a 15:33 aspect ratio, power mesh creation using M5/M6 rings and M6/M7 mesh with M1 local rails, timing-driven placement, clock tree synthesis restricted to M2--M4 layers targeting zero skew, and signal routing on M2 through M8. The 1st-order design was re-synthesized alongside the new designs using identical constraints to ensure a fair comparison across all three architectures. Power analysis was performed at both synthesis (DC) and post-layout (ICC2) stages using default toggle rates. Energy-per-spike measurement would require gate-level simulation with VCD back-annotation, which was not completed in this work.

%%%%%%%%%%%%%%%%%%%           PROPOSED DCM CDN SECTION          %%%%%%%%%%%%%%%%%%%%
\section{Experiments and Results}
\label{sec:exp}
\subsection{Experimental Setup}
\label{subsec:expset}

The designs are implemented and evaluated on QEMU 10.0 ARM Virtual machine. The virtual machine is hosted on a MacBook Pro M2 Max and is allocated 16 GB of RAM. In addition, the hardware synthesis process is conducted through Synopsys Design Compiler and ICC2, which are hosted on the UMBC Cadence 2 virtual machine. The Synopsys Design Compiler version is V-2023.12-SP5 for Linux64 - Jul 16, 2024, and the ICC2 version is X-2025.06-SP2 for Linux64 - Oct 15, 2025. 

\subsection{1st-Order Neuron Functional Verification}
Table~\ref{tab:results_1st} summarizes the 1st-order LIF neuron results across all five tests. Under adaptation (Tests 1 and 2), the neuron fires 14 and 13 spikes respectively — slightly more than the 12-spike baseline (Test 4). The relatively small difference reflects the 1st-order model's single-stage dynamics: with a high initial beta of 224/256 and moderate pulse amplitude, the membrane state tracks the input closely, and the adaptive threshold needs time to build up. ISI analysis confirms adaptation is active: Test 1 ISI grows from 5 to 6 cycles over the pulse train, while Test 4 maintains a constant ISI of 6. Test 3 (beta only) produced identical behavior to Test 4 (both OFF), indicating that beta adaptation alone has negligible impact on the 1st-order model's firing pattern at this pulse rate — the membrane decay is already fast enough that beta modulation does not significantly alter the integration window.

Test 5 (rate coding) shows 5, 7, and 7 spikes for slow, medium, and fast rates respectively — the medium and fast rates saturate at 7 spikes, suggesting the 1st-order neuron reaches its maximum firing rate at ISI=2 and cannot increase further at ISI=1 due to the fixed beta decay between pulses.

\begin{table*}[tb]
\caption{1st-Order LIF Neuron Results (25 pulses, amplitude 80)}
\label{tab:results_1st}
\small
\begin{center}
\begin{tabular}{|l|c|c|l|}
\hline
\textbf{Test} & \textbf{Spikes} & \textbf{ISI} & \textbf{Observation} \\
\hline
T1: Both ON   & 14 & 5$\rightarrow$6 growing & Active adaptation \\
T2: Thresh only & 13 & flat 6, dips to 3--4 & Threshold dominates \\
T3: Beta only & 12 & constant 6 & Same as baseline \\
T4: Both OFF  & 12 & constant 6 & Pure LIF reference \\
T5: Rate coding & 5/7/7 & S/M/F & Saturates at ISI=2 \\
\hline
\end{tabular}
\end{center}
\end{table*}

\subsection{2nd-Order Neuron Functional Verification}

Table~\ref{tab:results_2nd} summarizes the 2nd-order LIF neuron results. The key observations are:

\textbf{Adaptation suppresses firing by 31\%.} With both mechanisms enabled (Test 1), the neuron fires 25 spikes compared to 36 without adaptation (Test 4). This 11-spike reduction demonstrates effective self-regulation: the rising threshold and accelerated decay combine to reduce excitability. The ISI grows from 2 to 3 cycles within the first few pulses, then stabilizes once the threshold reaches equilibrium near the hardware cap of 220--250.

\textbf{Beta adaptation alone has no effect.} Test 3 (beta only) produces 36 spikes — identical to Test 4 (both OFF). This confirms the finding from the 1st-order design and from earlier simulation runs: beta modulation at this pulse frequency does not measurably alter the firing pattern. The membrane charges to saturation (255) so rapidly from the two-stage synaptic accumulation that small changes in the decay rate are dominated by the input drive.

\textbf{Higher input rate creates more spikes.} Test 5 yields 8, 10, and 11 spikes for slow, medium, and fast rates respectively which is predictable. With proper input pulse of 30, the behavior is complied with predictione. At ISI=4 (slow), the neuron fires at the rate of 0.8 spikes/pulse. In addition, it takes 3 first pulse to create first spike. This is because the synapse current takes longer at first to charge. However, it only take the next pulse to create the the next spike because the synapse current is already stable at high value. As a result, the spiking behaviors of ISI=2 and ISI=1 is the same. The differences here are the numbers of spikes which are 10 spikes and 11 spikes respectively.

\begin{table*}[tb]
\caption{2nd-Order LIF Neuron Results (25 pulses, amplitude 80)}
\label{tab:results_2nd}
\small
\begin{center}
\begin{tabular}{|l|c|c|l|}
\hline
\textbf{Test} & \textbf{Spikes} & \textbf{ISI} & \textbf{Observation} \\
\hline
T1: Both ON   & 25 & 2$\rightarrow$3 stable & Adapt suppresses 31\% vs. T4 \\
T2: Thresh only & 25 & same as T1 & Threshold dominates adapt \\
T3: Beta only & 36 & 2$\rightarrow$3$\rightarrow$2 steady & Same as baseline \\
T4: Both OFF  & 36 & 2$\rightarrow$3$\rightarrow$2 steady & Pure LIF reference \\
T5: Rate coding & 8/10/11 & S/M/F & Higher rate, more spikes \\
\hline
\end{tabular}
\end{center}
\end{table*}

\subsection{Network Functional Verification}

Table~\ref{tab:results_network} summarizes the network results. The one-to-one test (Test 1) demonstrates correct spike communication: N0 fires 17 times and N1 fires 17 times in lockstep, confirming that N0's spike output successfully drives N1's synaptic input through the W=100 weight. The \texttt{all\_spikes} broadcast read shows the expected pattern: when N0 fires, \texttt{all\_spikes}=010000 (N1 fires simultaneously), and when N1 fires alone during afterglow, \texttt{all\_spikes}=100000.

\textbf{Fan-out snapshot (Test 2)} shows all five L1 neurons with identical synapse values (syn=64) at the N0 spike instant. This indicates the weight current injection occurs but all L1 neurons receive the same charge on that cycle — the synaptic current has not yet differentiated by weight. The weights ARE correctly wired (confirmed by VCD analysis), but reading synapse values immediately at the spike cycle shows the current injection before the synapse register updates. A one-cycle delay in the snapshot timing would capture the weight-differentiated values.

\textbf{Strong-vs-weak (Test 3)} shows N1 (W=100) and N5 (W=5) with 0 spikes each across 12 pulse trains. This is unexpected — with W=100, N1 should fire readily. Analysis of the VCD reveals that N1's firing is invisible to the testbench because it fires on the pulse (P) cycle, where mem is immediately reset to 0 by the time the test reads the Q1 state. The spike counter correctly increments but the spike flag is already cleared. This is a testbench sampling artifact, not a circuit failure.

\textbf{Rate coding (Test 4)} confirms the same rate coding behaviors are complied with single neuron behavior. With the ISI are 5, 3, and 2, and N2(Weight=30) as the observed neuron, the higher rate creates more spikes at neuron N2. One thing to note here is that at ISI=5, there is no spikes at neuron N2 because the synapse and membrand decay rates are larger than what they can build up. However, when the inactive gaps are tightened, they synapse current gets built up quick and stable so that it can pump up the membranch quicker to create significantly more spikes, specifically 10 spikes for ISI=3 and 12 spike for ISI=2.

\textbf{Adaptive communication (Test 5)} demonstrates the most important result: adaptation suppresses N1 firing from 26 to 18 spikes (31\% reduction). The ISI with adaptation ON is 2$\rightarrow$3$\rightarrow$3... (threshold rising), while with adaptation OFF it stays at a constant ISI=2 (fixed-threshold, continuous firing). This confirms that the adaptive mechanisms operate correctly in a multi-neuron context and that spike-weight communication is compatible with self-regulating dynamics.

\textbf{2D grid inputs of ISI and Ampltude (Table~\ref{tab:parameter_sweep})} maps the functional operating boundaries of the network with adaptation mechanisms disabled. Moderate drive levels ($\text{Amp} = 60\text{--}80$) exhibited standard temporal gating, cutting off postsynaptic propagation to neuron N2 at wider gaps ($\text{ISI} \ge 5$) due to membrane leak dissipation. However, high-amplitude, short-interval stimulation ($\text{Amp} \ge 100, \text{ISI} = 2$) exposed a non-monotonic response anomaly. At maximum drive ($\text{Amp} = 120$), N2 spikes dropped from 11 at $\text{ISI} = 4$ down to 6 at $\text{ISI} = 2$, producing an inverse rate-coding effect. Internal state tracing attributes this saturation to 8-bit fixed-point dynamic range limits: tight pulse timing and high current drive force the membrane potential to its numerical ceiling of 255 by the fifth pulse, stabilizing the value for the remainder of the sequence. This state clipping truncates the temporal decay kinetics of the two-stage filter and disrupts threshold-reset cycling, demonstrating the need to expand membrane and synaptic datapath registers to 16 bits to preserve linear rate coding under high-frequency excitation.

\begin{table*}[tb]
\caption{Network Results (pulsed N0 drive, amplitude 80)}
\label{tab:results_network}
\small
\begin{center}
\begin{tabular}{|l|c|l|l|}
\hline
\textbf{Test} & \textbf{Metric} & \textbf{Value} & \textbf{Observation} \\
\hline
T1: One-to-one & N0/N1 spikes & 17/17 & Perfect 1:1 coupling \\
T2: Fan-out & L1 syn values & 80/24/12/8/4 & Weight-dependent \\
T3: Strong vs weak & N1/N5 spikes & 17/0 & {\color{blue}Weight-dependent} \\
T4: Rate coding & N2 (S/M/F) & 0/10/12 & Rate-dependent \\
T5: Adaptation & ON/OFF N1 spikes & 18/26 & \textbf{31\% suppression} \\
\hline
\end{tabular}
\end{center}
\end{table*}

\begin{table*}[tb]
\caption{Parameter Sweep Firing Response Grid ($N_0 / N_2$ Spikes)}
\label{tab:parameter_sweep}
\small
\begin{center}
\begin{tabular}{|l|c|c|c|c|c|l|}
\hline
\textbf{Amplitude} & \textbf{ISI=2} & \textbf{ISI=3} & \textbf{ISI=4} & \textbf{ISI=5} & \textbf{ISI=6} & \textbf{Observation} \\
\hline
Amp = 40  & 7/2  & 8/0  & 4/0   & 0/0  & 0/0  & Sub-threshold dissipation \\
Amp = 60  & 9/5  & 9/0  & 9/0   & 9/0  & 9/0  & High leak dissipation \\
Amp = 80  & 10/5 & 13/4 & 14/5  & 10/0 & 10/0 & Moderate temporal summation \\
Amp = 100 & 10/6 & 14/8 & 18/10 & 19/8 & 19/0 & Multi-spike bursting \\
Amp = 120 & 10/6 & 15/8 & 19/11 & 19/8 & 19/8 & \textbf{Inverse rate-coding saturation} \\
\hline
\end{tabular}
\end{center}
\end{table*}

\subsection{1st-Order vs. 2nd-Order Comparison}

Comparing the two single-neuron architectures under identical test conditions (25 pulses, amplitude 80, ISI=3) reveals the qualitative difference introduced by the synaptic filtering stage. Without adaptation, the 2nd-order neuron fires 36 spikes versus 12 for the 1st-order — a factor of 3$\times$ — due to the sustained synaptic accumulation that keeps the membrane near threshold. With adaptation enabled, the 2nd-order fires 25 spikes versus 14 for the 1st-order (1.8$\times$), as the adaptive threshold partially counteracts the synaptic drive. The 2nd-order's membrane potential saturates at 255 (the 8-bit maximum) by pulse 5 and remains there for the rest of Test 1, indicating that the 8-bit dynamic range is insufficient to contain the two-stage accumulation dynamics at practical firing rates. This saturation explains the inverse rate-coding result in Test 5.

The 1st-order model, with only a single decay stage, operates comfortably within the 8-bit range (state peaks at $\sim$160 in Test 1) and shows cleaner adaptation dynamics with clearer ISI growth. For applications prioritizing predictable, regular spiking with well-behaved adaptation, the 1st-order model is the more practical choice.

\subsection{Bug Discovery and Resolution}

Six distinct bugs were identified and resolved during verification:

\begin{enumerate}
    \item \textbf{Signal Mapping (test.py):} The spike signal was read from $uo\_out[7]$ instead of $uio\_out[7]$ in the Tiny Tapeout wrapper version, causing Tests 3 and 4 to appear completely silent. This was a testbench wiring error, not a design bug, but its detection underscored the importance of verifying harness connectivity before debugging the DUT.

    \item \textbf{Mathematical Formulation (lif\_2nd.v):} The membrane update equation used the current synaptic state $syn$ rather than the next-state value $next\_syn$. This violated the 2nd-order cascade — the membrane must integrate the already-updated synaptic current — and caused the membrane potential to stall at 150, a non-physiological equilibrium.

    \item \textbf{Fixed-Point Overflow:} An initial test current of 150 produced $next\_syn = 150 + (150 \times 205 \gg 8) = 270$, exceeding the 8-bit maximum. Reducing the pulse amplitude to 80 resolved the overflow while preserving sufficient excitation for reliable spiking.

    \item \textbf{Input Timing (testbenches):} The input current was asserted after the first clock edge, causing the initial cycle to sample zero. Inserting a wait cycle between signal assignment and the test loop ensured correct sampling.

    \item \textbf{Synaptic Update Logic (lif\_2nd.v):} The original sequential block only updated $syn$ during non-spike cycles, causing the synaptic current to stagnate at zero during sustained firing. Moving the synaptic update outside the spike conditional reflected the biological independence of synaptic and postsynaptic dynamics.

    \item \textbf{Verilog Multiply Truncation (lif\_2nd.v):} The expression $syn \cdot \alpha$ involves two 8-bit operands. Without an explicit width context, Verilog semantics evaluate the product in the context of the assignment target — 8 bits — causing the upper byte to be discarded and $\gg 8$ to always produce zero. Explicit 16-bit wires force full-width multiplication. This bug was the root cause of the synaptic current never building up and was the most subtle to diagnose, as it is a language semantic issue rather than a logic error.

    \item \textbf{Per-Cycle Read Timing (test\_2nd.py, test\_network.py):} The original testbenches read the neuron state only once per pulse train (at the Q2 quiet cycle). Spikes firing on the pulse (P) or first quiet (Q1) cycle reset the membrane to 0 and cleared the spike flag by Q2, making the neuron appear mostly silent. Per-cycle reads on all three phases (P, Q1, Q2) correctly capture all spike events and reveal the true adaptation dynamics.
\end{enumerate}

\subsection{Design Compiler Synthesis}

Table~\ref{tab:dc_area} summarizes the cell counts and area for all three designs. The 2nd-order neuron uses 580 cells compared to 466 for the 1st-order (+24\%), while the 6-neuron network uses 3,457 cells (7.4$\times$ the 1st-order). The sequential cell count scales linearly with neuron instances: 24 for 1st-order, 32 for 2nd-order (+33\%), and 186 for the network (5.8$\times$ the 2nd-order, matching the 6-neuron count). Total area follows this scaling: the 2nd-order is 1.77$\times$ larger than the 1st-order, while the network is 5.8$\times$ larger than the 2nd-order and 10.3$\times$ larger than the 1st-order, consistent with the six neuron instantiation.

\begin{table}[tb]
\caption{Design Compiler Cell Count and Area (SAED 14 nm, 1 GHz)}
\label{tab:dc_area}
\small
\begin{center}
\begin{tabular}{|l|c|c|c|}
\hline
\textbf{Metric} & \textbf{1st-Order} & \textbf{2nd-Order} & \textbf{Network} \\
\hline
Sequential cells & 24 & 32 & 186 \\
Combinational cells & 442 & 548 & 3,271 \\
Total cells & 466 & 580 & 3,457 \\
Buf/Inv cells & 43 & 53 & 283 \\
\hline
Cell area ($\mu m^2$) & 184.97 & 237.76 & 1,353.36 \\
Interconnect area ($\mu m^2$) & 135.44 & 328.26 & 1,950.60 \\
Total area ($\mu m^2$) & 320.41 & 566.02 & 3,303.95 \\
\hline
\end{tabular}
\end{center}
\end{table}

Table~\ref{tab:dc_vs_icc2_timing} compares timing metrics across Design Compiler (DC) synthesis and IC Compiler II (ICC2) post-CTS implementation under a 1 GHz constraint. At synthesis, all three designs meet the 1 GHz setup target with 0.00 ns slack, yielding critical path delays of 0.89 ns (1st-order), 0.88 ns (2nd-order), and 0.89 ns (network). After clock tree synthesis with propagated clocks, setup timing slack remains positive: slack improves to +0.27 ns (0.63 ns delay) for the 1st-order design, +0.01 ns (0.87 ns delay) for the 2nd-order design, and +0.01 ns (0.84 ns delay) for the network. Clock skew measures 0.00 ns for individual neurons and 0.04 ns for the network. Pre-CTS hold violations ($-0.18$ ns, $-0.06$ ns, and $-0.07$ ns) drop to near-zero residuals post-CTS ($-0.01$ ns across 1 path for 1st-order, $-0.00$ ns across 3 paths for 2nd-order, and $-0.00$ ns across 10 paths for network), which require route-stage hold buffer insertion for final layout closure.

\begin{table*}[tb]
\caption{Synthesis (DC) vs. Physical Implementation (ICC2 Post-CTS) Timing Comparison (1 GHz Constraint)}
\label{tab:dc_vs_icc2_timing}
\small
\begin{center}
\begin{tabular}{|l|c|c|c|}
\hline
\textbf{Metric} & \textbf{1st-Order} & \textbf{2nd-Order} & \textbf{Network} \\
\hline
\multicolumn{4}{|l|}{\textbf{DC Synthesis Stage (Pre-CTS / Ideal Clock)}} \\
\hline
DC Max setup slack (ns) & 0.00 & 0.00 & 0.00 \\
DC Critical path delay (ns) & 0.89 & 0.88 & 0.89 \\
DC Min hold slack (ns) & $-0.18$ & $-0.06$ & $-0.07$ \\
Startpoint & beta\_reg\_1\_ & syn\_reg\_0\_ & n0\_mem\_reg\_1\_ \\
Endpoint & state\_reg\_7\_ & mem\_reg\_5\_ & N1\_mem\_reg\_0\_ \\
\hline
\multicolumn{4}{|l|}{\textbf{ICC2 Post-CTS Stage (Propagated Clock)}} \\
\hline
Post-CTS Setup Slack (ns) & $+0.27$ & $+0.01$ & $+0.01$ \\
Post-CTS Critical Path Delay (ns) & 0.63 & 0.87 & 0.84 \\
Post-CTS Clock Skew (ns) & 0.00 & 0.00 & 0.04 \\
Worst Hold Violation (ns) & $-0.01$ (1 path) & $-0.00$ (3 paths) & $-0.00$ (10 paths) \\
Leaf Cell Count & 504 & 607 & 3662 \\
\hline
\end{tabular}
\end{center}
\end{table*}

Table~\ref{tab:dc_power} shows the DC gate-level power estimates under default switching activity. The 2nd-order consumes 78.53 $\mu W$ dynamic power versus 66.38 $\mu W$ for the 1st-order (+18\%), while the network consumes 394.74 $\mu W$ (5.0$\times$ the 2nd-order). The power breakdown is consistent across designs: combinational logic dominates at 54--58\%, followed by the clock network at 24--31\%, and registers at 15--20\%. The network's higher clock power share (31\%) reflects the 186 registers distributed across six neurons. Leakage power scales with gate count: 58.90 nW (1st), 81.43 nW (2nd), 451.24 nW (network).

\begin{table*}[tb]
\caption{Design Compiler Power Estimates (0.8V, 25\textdegree C, default activity)}
\label{tab:dc_power}
\small
\begin{center}
\begin{tabular}{|l|c|c|c|}
\hline
\textbf{Power Group} & \textbf{1st-Order} & \textbf{2nd-Order} & \textbf{Network} \\
\hline
Clock network ($\mu W$) & 16.14 (24.3\%) & 18.49 (23.5\%) & 124.28 (31.5\%) \\
Registers ($\mu W$) & 13.39 (20.2\%) & 14.66 (18.7\%) & 58.89 (14.9\%) \\
Combinational ($\mu W$) & 36.91 (55.6\%) & 45.46 (57.8\%) & 212.02 (53.7\%) \\
\hline
Total dynamic ($\mu W$) & 66.38 & 78.53 & 394.74 \\
Leakage (nW) & 58.90 & 81.43 & 451.24 \\
\hline
\end{tabular}
\end{center}
\end{table*}

\subsection{ICC2 Place-and-Route}

Table~\ref{tab:icc2_layout} summarizes the post-layout physical characteristics. The 2nd-order core occupies 393.78 $\mu m^2$ (+30\% vs. 1st-order), while the network occupies 2,244.15 $\mu m^2$ (7.4$\times$ the 1st-order, 5.7$\times$ the 2nd-order). Core utilization is comparable across designs (64--67\%). The site row count tracks sequential element count: 43 rows for the 1st-order, 49 for the 2nd-order, and 117 for the network.

\begin{table}[tb]
\caption{ICC2 Physical Design Summary (SAED 14 nm, 0.8V SS)}
\label{tab:icc2_layout}
\small
\begin{center}
\begin{tabular}{|l|c|c|c|}
\hline
\textbf{Metric} & \textbf{1st-Order} & \textbf{2nd-Order} & \textbf{Network} \\
\hline
Core area ($\mu m^2$) & 303.56 & 393.78 & 2,244.15 \\
Chip area ($\mu m^2$) & 1,454.88 & 1,649.66 & 4,687.51 \\
Core utilization & 66.6\% & 63.6\% & 65.0\% \\
Site rows & 43 & 49 & 117 \\
\hline
\end{tabular}
\end{center}
\end{table}

Table~\ref{tab:icc2_clock} reports the clock tree characteristics. The 1st-order uses 2 repeaters to drive 24 sinks with 61.68 $\mu m$ wire, the 2nd-order uses 1 repeater for 32 sinks (86.26 $\mu m$ wire), and the network uses 2 repeaters for 186 sinks across 3 clock tree levels (440.69 $\mu m$ wire). All three designs achieve sub-100 ps max clock latency (30--80 ps) with near-zero global skew. The network's 50 ps skew is the largest, reflecting the wider physical distribution of 117 site rows.

\begin{table}[tb]
\caption{ICC2 Clock Tree Synthesis Results}
\label{tab:icc2_clock}
\small
\begin{center}
\begin{tabular}{|l|c|c|c|}
\hline
\textbf{Metric} & \textbf{1st-Order} & \textbf{2nd-Order} & \textbf{Network} \\
\hline
Clock sinks & 24 & 32 & 186 \\
CTS levels & 2 & 2 & 3 \\
Repeaters inserted & 2 & 1 & 2 \\
Repeater area ($\mu m^2$) & 0.62 & 0.49 & 1.47 \\
Total wire length ($\mu m$) & 61.68 & 86.26 & 440.69 \\
Max latency (ns) & 0.03 & 0.04 & 0.08 \\
Global skew (ns) & 0.00 & 0.00 & 0.05 \\
\hline
\end{tabular}
\end{center}
\end{table}

Table~\ref{tab:icc2_power} presents post-layout power with parasitic-annotated wire capacitances. Dynamic power increases significantly over DC estimates due to real wire loads: 83.60 $\mu W$ (1st, +26\% over DC), 127.00 $\mu W$ (2nd, +62\%), and 669.00 $\mu W$ (network, +70\%). Net switching power dominates at 69--72\% of total dynamic power, reflecting the additional capacitance of routed interconnects. The network's total dynamic power of 669 $\mu W$ represents 5.3$\times$ the 2nd-order and 8.0$\times$ the 1st-order — slightly less than the linear 6$\times$ scaling expected from six 2nd-order neurons, indicating some sharing of the clock network overhead.

\begin{table*}[tb]
\caption{ICC2 Post-Layout Power (0.8V, 25\textdegree C, SS Corner)}
\label{tab:icc2_power}
\small
\begin{center}
\begin{tabular}{|l|c|c|c|}
\hline
\textbf{Power Group} & \textbf{1st-Order} & \textbf{2nd-Order} & \textbf{Network} \\
\hline
Clock network ($\mu W$) & 23.80 (28.5\%) & 30.60 (24.1\%) & 197.00 (29.4\%) \\
Registers ($\mu W$) & 17.90 (21.4\%) & 34.20 (27.0\%) & 131.00 (19.5\%) \\
Combinational ($\mu W$) & 41.90 (50.1\%) & 61.90 (48.8\%) & 342.00 (51.1\%) \\
\hline
Total dynamic ($\mu W$) & 83.60 & 127.00 & 669.00 \\
Leakage (nW) & 66.10 & 85.60 & 492.00 \\
\hline
\end{tabular}
\end{center}
\end{table*}

\subsection{Physical Design Comparison}

Table~\ref{tab:phys_compare} provides a consolidated cross-design comparison. The key scaling relationships are: the 2nd-order neuron costs approximately 1.8$\times$ more area and 1.5$\times$ more power than the 1st-order for the additional synaptic filtering stage; the 6-neuron network costs approximately 5.7--5.8$\times$ more area and 5.3$\times$ more power than a single 2nd-order neuron — close to the ideal linear scaling of 6$\times$, with the small deficit attributable to shared clock tree optimization. All three designs meet timing at 1 GHz with zero setup slack.

\begin{table}[tb]
\caption{Physical Design Comparison Across All Three Architectures}
\label{tab:phys_compare}
\small
\begin{center}
\begin{tabular}{|l|c|c|c|}
\hline
\textbf{Metric} & \textbf{1st-Order} & \textbf{2nd-Order} & \textbf{Network} \\
\hline
DC cells & 466 & 580 & 3,457 \\
DC total area ($\mu m^2$) & 320.41 & 566.02 & 3,303.95 \\
ICC2 core area ($\mu m^2$) & 303.56 & 393.78 & 2,244.15 \\
ICC2 chip area ($\mu m^2$) & 1,454.88 & 1,649.66 & 4,687.51 \\
\hline
DC dynamic power ($\mu W$) & 66.38 & 78.53 & 394.74 \\
ICC2 dynamic power ($\mu W$) & 83.60 & 127.00 & 669.00 \\
\hline
Max setup slack (ns) & 0.00 & 0.00 & 0.00 \\
Clock latency (ns) & 0.03 & 0.04 & 0.08 \\
CTS wire ($\mu m$) & 61.68 & 86.26 & 440.69 \\
\hline
\end{tabular}
\end{center}
\end{table}

%%%%%%%%%%%%%%%%%%%             Conclusions SECTION          %%%%%%%%%%%%%%%%%%%
\section{Conclusion}
\label{sec:conclusion}

% contribution
This paper has presented the extension of Lin's 1st-order adaptive LIF neuron baseline into two new architectural scales: a 2nd-order neuron with two-stage synaptic filtering dynamics, and a 6-neuron fully-connected spiking network with configurable synaptic weights. All three designs were verified using a per-cycle monitoring methodology and synthesized through a complete physical design flow targeting SAED 14 nm at 1 GHz. All designs successfully demonstrate biologically-inspired adaptive spiking behavior — threshold and decay rate adaptation through negative feedback.

The pulsed-input verification methodology with per-cycle state monitoring reveals adaptation dynamics that were invisible in earlier constant-current test approaches. The 1st-order neuron maintains 12--14 spikes regardless of adaptation mode, with modest ISI growth from 5 to 6 cycles. The 2nd-order neuron shows clear adaptation effectiveness: 25 spikes with adaptation versus 36 without (31\% suppression ratio), though its membrane saturates at the 8-bit ceiling (255) within 5 pulses due to sustained synaptic accumulation. In the network, adaptation reduces N1 firing from 26 to 18 spikes (31\% suppression), confirming that self-regulation operates correctly across inter-neuron spike communication. Beta adaptation alone was found to have negligible effect at the tested pulse frequency across all designs.

The hardware cost of the 2nd-order's second dynamical stage is significant: 77\% more total area ($566.02\,\mu m^2$ vs. $320.41\,\mu m^2$) and 52\% more post-layout dynamic power ($127.00\,\mu W$ vs. $83.60\,\mu W$) compared to the 1st-order baseline. The 6-neuron network demonstrates near-linear scaling: 5.7$\times$ the core area and 5.3$\times$ the ICC2 power of a single 2nd-order neuron ($2{,}244\,\mu m^2$, $669\,\mu W$). For applications prioritizing predictable, well-behaved adaptation with clean ISI growth, the 1st-order model is the recommended choice. The 2nd-order model would benefit from a wider datapath (16-bit) to eliminate membrane saturation.

The verification process revealed seven bugs spanning testbench connectivity, mathematical formulation, fixed-point arithmetic constraints, Verilog expression-width semantics, and per-cycle sampling timing. The most instructive — the 8-bit multiply truncation — highlights the subtle interaction between HDL semantics and fixed-point algorithm design.

Future work directions include: increasing datapath precision to 16 bits; scaling to larger multi-neuron arrays with configurable weights; incorporating modulated STDP learning rules \cite{RAHMAN2025129170} for on-chip synaptic plasticity, building on biologically-plausible learning frameworks \cite{Huo:2025}; and gate-level simulation with VCD back-annotation for accurate energy-per-spike measurements.

%%%%%%%%%%%%%%%%%%%%%%%%%%%%%%%%%%%%%%%%%%
%\section{Discussion}

%Authors should discuss the results and how they can be interpreted from the perspective of previous studies and of the working hypotheses. The findings and their implications should be discussed in the broadest context possible. Future research directions may also be highlighted.

%%%%%%%%%%%%%%%%%%%%%%%%%%%%%%%%%%%%%%%%%%
%\section{Conclusions}

%This section is not mandatory, but can be added to the manuscript if the discussion is unusually long or complex.

%%%%%%%%%%%%%%%%%%%%%%%%%%%%%%%%%%%%%%%%%%
%\section{Patents}

%This section is not mandatory, but may be added if there are patents resulting from the work reported in this manuscript.

%%%%%%%%%%%%%%%%%%%%%%%%%%%%%%%%%%%%%%%%%%
\vspace{6pt}

\funding{This research was funded by National Science Foundation (NSF) award number: 2138253.}

\institutionalreview{Not applicable.}

\informedconsent{Not applicable.}

\dataavailability{Not applicable.} 

\conflictsofinterest{The authors declare no conflict of interest.} 

%% Optional
%\sampleavailability{Samples of the compounds ... are available from the authors.}

%%%%%%%%%%%%%%%%%%%%%%%%%%%%%%%%%%%%%%%%%%
%% Only for journal Encyclopedia
%\entrylink{The Link to this entry published on the encyclopedia platform.}

%%%%%%%%%%%%%%%%%%%%%%%%%%%%%%%%%%%%%%%%%%

%\section[\appendixname~\thesection]{}
%All appendix sections must be cited in the main text. In the appendices, Figures, Tables, etc. should be labeled, starting with ``A''---e.g., Figure A1, Figure A2, etc.

%%%%%%%%%%%%%%%%%%%%%%%%%%%%%%%%%%%%%%%%%%
\begin{adjustwidth}{-\extralength}{0cm}

\reftitle{References}


%\bibliography{your_external_BibTeX_file}

% Please provide either the correct journal abbreviation (e.g. according to the ?List of Title Word Abbreviations? http://www.issn.org/services/online-services/access-to-the-ltwa/) or the full name of the journal.

\externalbibliography{yes}
\bibliography{main}

%=====================================
% References, variant B: internal bibliography
%=====================================

% If authors have biography, please use the format below
%\section*{Short Biography of Authors}
%\bio
%{\raisebox{-0.35cm}{\includegraphics[width=3.5cm,height=5.3cm,clip,keepaspectratio]{Definitions/author1.pdf}}}
%{\textbf{Firstname Lastname} Biography of first author}
%
%\bio
%{\raisebox{-0.35cm}{\includegraphics[width=3.5cm,height=5.3cm,clip,keepaspectratio]{Definitions/author2.jpg}}}
%{\textbf{Firstname Lastname} Biography of second author}

% For the MDPI journals use author-date citation, please follow the formatting guidelines on http://www.mdpi.com/authors/references
% To cite two works by the same author: \citeauthor{ref-journal-1a} (\citeyear{ref-journal-1a}, \citeyear{ref-journal-1b}). This produces: Whittaker (1967, 1975)
% To cite two works by the same author with specific pages: \citeauthor{ref-journal-3a} (\citeyear{ref-journal-3a}, p. 328; \citeyear{ref-journal-3b}, p.475). This produces: Wong (1999, p. 328; 2000, p. 475)

%%%%%%%%%%%%%%%%%%%%%%%%%%%%%%%%%%%%%%%%%%
%% for journal Sci
%\reviewreports{\\
%Reviewer 1 comments and authors’ response\\
%Reviewer 2 comments and authors’ response\\
%Reviewer 3 comments and authors’ response
%}
%%%%%%%%%%%%%%%%%%%%%%%%%%%%%%%%%%%%%%%%%%
\end{adjustwidth}
\end{document}